\documentclass{article}

\PassOptionsToPackage{numbers}{natbib}
\usepackage[final]{neurips_2021}

\usepackage[utf8]{inputenc} 
\usepackage[T1]{fontenc}    
\usepackage{hyperref}       
\usepackage{url}            
\usepackage{booktabs}       
\usepackage{amsfonts}       
\usepackage{nicefrac}       
\usepackage{microtype}      
\usepackage{xcolor}         

\usepackage{algorithm}
\usepackage[linesnumbered,ruled,algo2e]{algorithm2e}
\usepackage{graphicx}
\usepackage{amsthm,amsmath,amssymb,bbm}
\usepackage{mathtools}
\usepackage{forest}
\usepackage{subfigure}

\title{A Validation Tool for Designing Reinforcement Learning Environments}

\author{%
  Ruiyang Xu \\
  Northeastern University\\
  Boston, MA 02115 \\
  \texttt{ruiyang@ccs.neu.edu} \\
   \And
   Zhengxing Chen \\
   Facebook \\
   Menlo Park, CA 94025 \\
   \texttt{czxttkl@fb.com} \\
}

\begin{document}

\maketitle

\begin{abstract}
  Reinforcement learning (RL) has gained increasing attraction in the academia and tech industry with launches to a variety of impactful applications and products. Although research is being actively conducted on many fronts (e.g., offline RL, performance, etc.), many RL practitioners face a challenge that has been largely ignored: determine whether a designed Markov Decision Process (MDP) is valid and meaningful. This study proposes a heuristic-based feature analysis method to validate whether an MDP is well formulated. We believe an MDP suitable for applying RL should contain a set of state features that are both sensitive to actions and predictive in rewards. We tested our method in constructed environments showing that our approach can identify certain invalid environment formulations. As far as we know, performing validity analysis for RL problem formulation is a novel direction. We envision that our tool will serve as a motivational example to help practitioners apply RL in real-world problems more easily.
\end{abstract}

\section{Introduction}
Reinforcement Learning (RL) aims to learn a policy to make sequential decisions for optimizing long-term rewards. It has been studied for decades~\cite{Sutton1998} and recently got increasing attention thanks to break-through applications in board games~\cite{silver2016mastering} and video games~\cite{vinyals2019grandmaster,berner2019dota}. Ever since, researchers have pushed the frontier of RL learning performance~\cite{haarnoja2018soft,hafner2020mastering,mnih2013playing,fujimoto2018addressing}, tools, and platforms~\cite{castro2018dopamine,liang2018rllib,gauci2018horizon}. Practitioners have also applied RL to a wide range of real-world applications and products~\cite{li2019reinforcement}, e.g., device placement~\cite{mirhoseini2017device}, recommendation systems~\cite{zhao2018deep,chen2019top}, and ridesharing~\cite{xu2018large}.



While most existing works deal with environments with a well-designed Markov Decision Process (MDP), RL practitioners need to carefully design a valid MDP to carry out meaningful and effective optimization when it comes to a new domain. Without a validation tool, it would be hard to understand if they can successfully apply RL algorithms to optimize long-term value based on the designed action and state space.

Let us take email marketing as an example. Suppose we would like to learn a policy for personalizing campaign emails to users. We design an action space that includes different personalization levers, such as email styles, delivery time, and promotion types. We can observe a set of user-side features as the state and regard users' purchase value as the reward. Should we apply contextual bandits or sequential decision RL algorithms? Are the personalization levers effective in navigating the user features into promising regions? Do the included user features really help us differentiate user preferences in the purchase? The challenge of these questions can be magnified when the RL practitioner works at a large-scale corporate where it is hard to know every detail of data because feature engineering is distributed to different teams. Although the best form of problem formulation may be identified through extensive experiments in an online A/B test system, it would be time-consuming and costly.

We assume that a well-formed MDP should satisfy the following properties: 

\textbf{Property 1:} There exist some state features that are predictive of the reward. We call such features \textit{reward-contributing features}. If there is no reward-contributing state feature, the design of the state space is futile. The problem is then a bandit problem at best. 

\textbf{Property 2:} Among the reward-contributing features, at least one of them should also be \textit{action-sensitive}, i.e., sensitive to the change of actions. If no reward-contributing feature is sensitive to actions, then the agent cannot control the state transition towards the regions that are promising for value optimization. 

This paper proposes a model-based feature analysis method for detecting if an MDP is formulated in a valid and meaningful way. We will train a model to learn the statistical dependency between the state, action, and reward in the designed MDP. By probing the model with a simple causal principle~\cite{reichenbach1956direction}, we determine whether Property 1 and 2 are satisfied.

Our work is premised on causal inference~\cite{pearl2009causality,peters2017elements}, a broad topic which has helped RL research in different ways, such as counterfactual policy evaluation~\cite{rezende2020causally,buesing2018woulda} and model evaluation~\cite{bottou2013counterfactual}. However, the challenge of validating RL problem formulation has been largely ignored. To our best knowledge, we only know a related work from~\cite{shi2020does}, which proposed a methodology for testing if \textit{Markov Assumption} holds for an environment (i.e., the optimal policy can make decisions solely based on the most recent state and action without the need to memorize historical experience). The validation provided by our tool is testing different aspects (Property 1 and 2) and thus considered complementary to the tests in~\cite{shi2020does}.

\section{Methodology}
\subsection{Markov Decision Process}
Reinforcement learning focuses on solving sequential decision problems under the framework of Markov Decision Processes (MDPs). An MDP contains the following components:
(1) State Space $\mathcal{S}$, which contains all possible states $s\in \mathcal{S}$ in a decision problem. We denote the dimension of $\mathcal{S}$ as $d$, i.e., $s=[f_1,f_2,...,f_d]$. (2) Action Space $\mathcal{A}$, which contains all possible actions  $a\in \mathcal{A}$ in a decision problem. (3) Transition probabilities $\mathcal{T}(s'|s,a)$, which defines the dynamic from one state to another, namely, taking action $a$ at state $s$ has a probability $\mathcal{T}(s'|s,a)$ to arrive at state $s'$. (4) Rewards $\mathcal{R}(s,a,s')$, which defines the expected reward after taking action $a$ in state $s$ and moving to state $s'$. (5) Reward discount factor $\gamma\in (0,1]$, which weighs the importance of future rewards. 



Decisions making on a specific MDP can be abstracted as a policy $\pi(a|s)$, which defines the probability distribution of taking some action $a$ given some state $s$. Solving an MDP means to find an optimal policy that maximizes the 
state value function:
$V^{\pi}(s)=\mathbb{E}_{\pi}[ \,G_t|s_t=s]=\mathbb{E}_{\pi}[\sum_{i=0}^{T-t}\gamma^i R(s_{t+i},a_{t+i},s_{t+i+1}) | s_t=s]$,
where $G_t$ denotes the cumulative discounted reward and $T$ is the maximal episode's length.


\subsection{World Model}
	
Our method is based on the world model \cite{ha2018recurrent}, a deep learning model for modeling the state transition and reward function. The world model uses a Mixture Density Recursive Neural Network (MDN-RNN) as a powerful representation for the state transition and reward function. The MDN-RNN takes in the last $M$-step states and actions while predicting the next state and reward. Following the world model work~\cite{ha2018recurrent}, the reward is predicted as a pointwise scalar, whereas the next state is represented as a mixture Gaussian density - the network learns the mean, standard deviation, and affinity for each multivariate Gaussian (with diagonal covariance)~\cite{Bishop94mixturedensity}. We believe our methodology is general enough that other advanced modeling techniques can work as well (See~\cite{moerland2020model} for an overview).

In this research, we only care about predicting the reward and the next state using the current state and action (i.e., $M=1$). Therefore, we use the following loss function to train the RNN:
$$L= \underbrace{(\hat{r}_t(s_t, a_t) - r_{t})^2}_\text{reward loss}\underbrace{-log\left[\sum_{k=1}^K\alpha_k(s_t,a_t)\mathcal{N}(s_{t+1}|\mu_k(s_t,a_t),\sigma_k(s_t,a_t))\right]}_\text{transition loss},$$
where $r_t$ and $s_{t+1}$ are the target reward and target next state. $\hat{r}_t$ is the predicted reward. $\alpha_k$ is the predicted affinity to the $k$-th Gaussian in the mixture, while $\mu_k$ and $\sigma_k$ are the predicted mean and standard deviation for the $k$-th Gaussian from the output heads of the RNN.




\section{Property Verification}
We first train a group of $N$ world models with different initialization seeds on the training split of a given RL dataset $\{(s_t, a_t, r_t, s_{t+1})\}_{t=1}^B$. The dataset is collected by a random policy to ensure no confounder between actions and states. With the ensemble of the trained world models, we apply feature analysis on the evaluation split based on the following metrics:
\begin{itemize}
    \item Reward contribution: a state feature $f_i$'s contribution to reward prediction is measured by a perturbation-based feature importance method. We look at the increase of Mean Absolute Error (MAE) for reward prediction after replacing all the values of $f_i$ in the evaluation data with its mean value. The more MAE increases, the more critical a state feature is to predict the reward. In practice, we accumulate reward contribution by mini-batches; the mean feature values are computed and set within each mini-batches instead of over the whole evaluation dataset. 
    \item Action sensitivity: the action sensitivity of a state feature $f_i$, is measured by the change of the next state prediction after we randomly shuffle the input actions within mini-batches of the evaluation data. Low sensitivity indicates that the actions cannot control $f_i$'s state transition effectively. 
\end{itemize}

\begin{figure}[t]
\includegraphics[width=\linewidth]{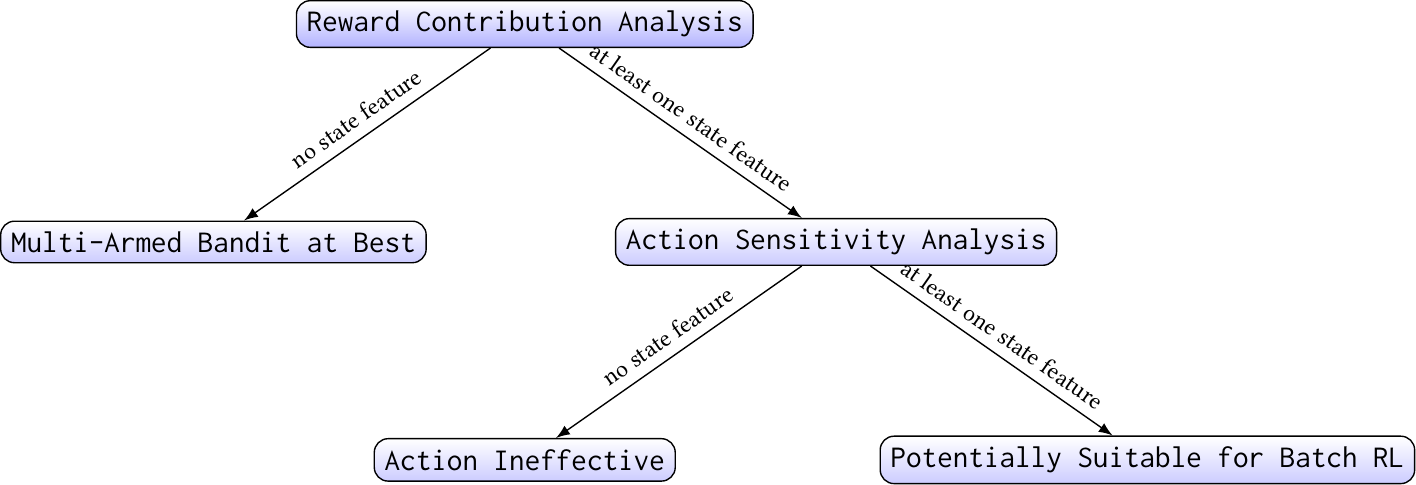}
\caption{The process of feature analysis}
\label{fig:5}
\end{figure}

Since we use MDN-RNN as the world model, the prediction for the next states is a Gaussian mixture. We design a formula to quantify the action sensitivity based on the Gaussian mixture output:
    \begin{equation*}
    \begin{split}
    \begin{aligned}
        &\textbf{Action\_ Sensitivity}(s)=\sum\limits_{i=1}^{|\mathcal{D}|}\sum\limits_{k=1}^K \frac{\alpha_k(s_i,a_i) + \alpha_k(s_i,\bar{a}_i)}{2} \cdot |\mu_k(s_i, a_i) - \mu_k(s_i, \bar{a}_i)|,
    \end{aligned}
    \end{split}
    \end{equation*}
where $\bar{a}_i$ denotes shuffled actions. The intuition behind this formula is that we measure the difference of the predicted means weighted by $(\alpha_k(s_i,a_i) + \alpha_k(s_i,\bar{a}_i))/2$, which is the average affinity to the $k$-th Gaussian resulted from the original and shuffled input. Therefore, if a Gaussian in the mixture has little weight by both the original and shuffled results, the difference from that mixture would be moderated as well in the overall action sensitivity.

Our feature analysis is rooted in \textbf{Reichenbach's common causal principle}~\cite{reichenbach1956direction}, which states that if two random variables $X$ and $Y$ are statistically dependent, then there exists a third variable $Z$ that causally influences both. (As a special case, $Z$ may coincide with $X$ or $Y$.) In our method, statistical dependence is learned through the world model, while we make some important assumptions to eliminate unrealistic causal relationships. The assumptions we make are: (1) actions may or may not cause the change of rewards or states, but states or rewards cannot cause the change of actions; (2) states may or may not cause the change of rewards, but rewards cannot cause the change of states. Since our data is collected by a random policy, we also assume that the possibility that action and state dependence learned by the world model (if there is any) is \textit{not} due to a confounder between them.

In reward contribution analysis, shuffling little reward-contributing features are unlikely to reduce the reward prediction MAE. However, performing action sensitivity analysis will return a non-negative number for any feature, since our analysis examines the \textit{magnitude} of changes of deep learning models' predictions. We need a baseline to quantitatively filter out little action-sensitive features. For this purpose, we design a simple method to decide the filtering thresholds. We train another ensemble of $N$ world models with the exact same training dataset, except that the actions in the data are randomly shuffled (within mini-batches). The learned world models on shuffled actions are expected to be meaningless, but its action sensitivity level will be used as a baseline to offset the actual action sensitivity result. 

We only consider results being significant when the $X$-percentile of the reward contribution or \textit{offset} action sensitivity is above 0, where $X$ can be set to a global value ($75\%$ in this paper) as default or a per feature-based threshold if the user has more prior domain knowledge. For example, if the user knows that one state feature is sensitive to the change of actions only under certain situations (e.g., in 10\% of all encountered states), then setting $X\%=75\%$ would be too strict. The population of a feature's reward contribution or action sensitivity statistics is of {\fontfamily{lmtt}\selectfont $N \cdot num\_eval\_batches$} size, from which $X$-percentiles are computed.

In practice, when there are multiple correlated state features, we also notice that deep learning models may focus on a subset of them instead of giving them equal importance. This will be problematic in the reward contribution analysis because we may erroneously conclude that only one feature effectively predicts rewards, while other correlated state features are also compelling. To combat this issue, we apply random dropout \textit{on the input state features}~\cite{volkovs2017dropoutnet,srivastava2014dropout} such that world models will not overly utilize one among all correlated features.

The overall flow of our feature analysis methodology can be summarized in Figure~\ref{fig:5}. We first test the existence of reward-contributing state features. If there is no such feature, then we conclude that the given dataset is \textit{not} suitable for RL. At best, one might want to try other approaches like Bandit algorithms. On the other hand, if at least one state feature is reward-contributing, we check whether it is action-sensitive. If none is action-sensitive among all reward-contributing state features, then it indicates that actions have no control over state transitions. We can only conclude that the MDP formulation is \textit{potentially} suitable for RL when there is a non-empty state feature set identified as both reward-contributing and action-sensitive.

\subsection{Limitations}\label{sec:limitations}
	
There are three major limitations to our current methodology. First, we assume we can access a dataset collected by randomized actions, which assures us that if any state is identified as action-sensitive, there is a causal link between actions to the state. This assumption may not be feasible for certain offline batch RL scenarios, where we are only given a dataset without prior information on the logging policy. Second, even though there is no confounder between actions and states, there may exist confounders between some states and rewards such that even those states are identified as reward-contributing, they cannot \textit{cause} the change of rewards. In this case, it would still be futile to apply RL algorithms to optimize returns. Third, our methodology is largely heuristics-driven, which may not perform as expected in corner cases. For example, our action sensitivity formula does not take into accounts the predicted standard deviations of Gaussian mixtures. We would erroneously conclude that a state is not action-sensitive if actions can only change the state's variance but not its mean value.

\section{Experiment}
\subsection{Environments Specifications}\label{sec:env_fab}
We constructed different environments to simulate different causal relations between states, actions, rewards, and possible hidden factors. All the environments generate episodes of data of $T=10$ steps. All the environments have the same dimension $d=10$ of the state space yet different environments have varied state transition $\mathcal{T}(s'|s,a)$ and reward functions $\mathcal{R}(s,a,s')$. For proof of the concept, there are only two possible discrete actions, 0 or 1. The training batches for world models and the evaluation batches used for feature analysis will be sampled by a random policy (i.e., tossing a coin). In our feature analysis, the ensemble size $N=10$ applies to the original group of world models and the baseline group of world models for action sensitivity. We deliberately set $K=5$ (the number of Gaussian mixture components), which is higher than the expected number of possible stochastic transitions in our experiment design - in our designed environments, current states could have at most two possible next states. However, we overparameterize $K$ to show that results can be robust even though the user does not know prior information about state transitions and opts to set $K$ to a high value for being more flexible. Please refer to Table~\ref{tab:hyperparam} in Appendix for all hyper-parameters used in the experiments. 

Each constructed environment as well as our expected outcomes are described as follows. Their causal relationships are shown in Figure~\ref{result_plot} in Appendix.
\begin{enumerate}
    \item Null relation: in this case, none of the state, action, and reward is dependent on each other. The state transition is a stochastic "adding one" process,  defined as:
    $$P(f_i^{t+1}=f_i^t+1)=1-\frac{i}{d}, \qquad P(f_i^{t+1}=f_i^t)=\frac{i}{d}$$
    
    The reward is a random signal from either 0 or 1 with a probability of 0.5. The environment works as a baseline, which should return no significant results because everything is independent. 

    \item Action to reward causal relation: everything is kept independent as in (2) except that the actions have a causal relation with the reward; $P(r=1|a=1)=1$.

    \item Action to state causal relation: As in (1) except that the state will change with different probabilities depending on which action to take: $$P(f_i^{t+1}=f_i^t+1|a=1)=1-\frac{i}{d}, \quad P(f_i^{t+1}=f_i^t|a=1)=\frac{i}{d}, \quad P(f_i^{t+1}=f_i^t|a=0)=1$$

    \item State to reward causal relation: only states have a causal relation with the rewards; the reward signal will change with the probability $P(r=1|f_0^t>4)=1$, i.e., the reward will be one only if one specific feature reaches more than 4.

    \item Action to the state to reward causal relation: in this case, we have a full causal relation, which means that the state transition behaves like (3), while the reward acts like (4).

    \item Hidden confounder to state and reward. In this case, we have conditionally independent states and rewards, given a hidden factor $h$'s, which take value either 0 or 1. The hidden factor $h$ is randomly initialized at the beginning of an episode and then kept fixed until the episode terminates. The state will change with the probabilities, which are independent of actions:
    $$P(f_i^{t+1}=f_i^t+1|h=1)=1-\frac{i}{d}, \quad P(f_i^{t+1}=f_i^t|h=1)=\frac{i}{d}$$
    $$P(f_i^{t+1}=f_i^t+1|h=0)=\frac{i}{d}, \quad P(f_i^{t+1}=f_i^t|h=0)=1 - \frac{i}{d}$$
    Meanwhile the reward changes with the probability:
    $$P(r=1|h=1)=0.8, \quad P(r=0|h=1)=0.2$$
    $$P(r=1|h=0)=0.2, \quad P(r=0|h=1)=0.8$$
    
    We would expect no action-sensitive states but reward contributing states identified by our methodology because of the deliberately introduced confounder between states and rewards.

    \item Action to state with a hidden factor as a confounder to state and reward. In this case, like in (6), the hidden factor affects states and rewards, but the state is also affected by the actions chosen. As a result, the state changes with the following probabilities:
    $$P(f_i^{t+1}=f_i^t+1|h=1,a=1)=1-\frac{i}{d}, \quad P(f_i^{t+1}=f_i^t|h=1,a=1)=\frac{i}{d}$$
    $$P(f_i^{t+1}=f_i^t|h=1,a=0)=1, \quad P(f_i^{t+1}=f_i^t+1|h=0,a=1)=\frac{i}{d}$$
    $$P(f_i^{t+1}=f_i^t|h=0,a=1)=1-\frac{i}{d}, \quad P(f_i^{t+1}=f_i^t|h=0,a=0)=1$$
    
    In this case, we would expect that our methodology identifies all states as both action-sensitive and reward-contributing.
\end{enumerate}

\subsection{Experimental Results}
Reward contribution and action sensitivity for each environment are reported in Figure~\ref{result_plot} in Appendix. The box plots use the lower edge of boxes to denote the 75\%-percentile. Therefore, we can visually identify if results are significant. We can see that all results meet our expectations. In other words, for states we expect to have statistically significant reward contribution, we indeed find their 75\% percentile reward contribution is above 0; for states expected to be action sensitive, we also find their 75\% percentile action sensitivity adjusted by the baseline ensemble to be above 0.

\section{Conclusion}
	
We proposed a feature analysis method to help RL practitioners evaluate the appropriateness of MDP formulation and suitability for applying RL algorithms. Our approach can be used to accelerate feature engineering iterations and potentially improve training performance. RL algorithms are suitable for problems: (1) taking actions can effectively lead to state transitions, (2) rewards are predictable by states or actions. If either condition is unsatisfied, it raises a flag for a more in-depth understanding of data. The world model is at the core of our method, which simulates the underlying MDP from the logged data. We perform experiments on constructed environments with various causal relationships between states, actions, and rewards. The results are consistent with the environment design and our expectations. Our methodology comes with several limitations. It is mainly heuristic-driven; thus, it heavily depends on data and model quality. We need to make a few assumptions (no confounder between actions and states or between states and rewards) in order to make our results valid.

\bibliographystyle{ACM-Reference-Format} 
\bibliography{main}
\appendix
\section{Appendix}

\begin{table}[h]
\centering
  \caption{Hyperparameters used in the experiments}
  \label{tab:hyperparam}
  \begin{tabular}{cc}
    \toprule
    Parameter & Value \\
    \midrule
    {\fontfamily{lmtt}\selectfont num\_train\_batches} & 1000\\
    {\fontfamily{lmtt}\selectfont num\_eval\_batches} &  200\\
    $d$ (state dimension) & 10 \\
    {\fontfamily{lmtt}\selectfont num\_of\_actions} &  2\\
    $T$ (maximal episode length) & 10 \\
    $N$ (world model ensemble size) & 10 \\
    $K$ (MDN-RNN \# of Gaussians) & 5 \\
    MDN-RNN \# of hidden layers & 2 \\
    MDN-RNN hidden layer size & 32 \\
    mini-batch size & 1024 \\
  \bottomrule
\end{tabular}
\end{table}

\begin{figure}[h]
      \centering
      \subfigure[]{


\includegraphics[width=0.2\linewidth]{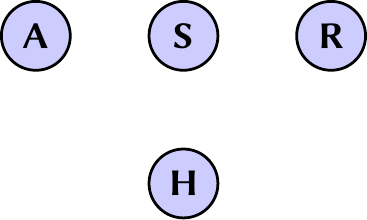}
      }
      \subfigure[]{\includegraphics[width=0.3\linewidth]{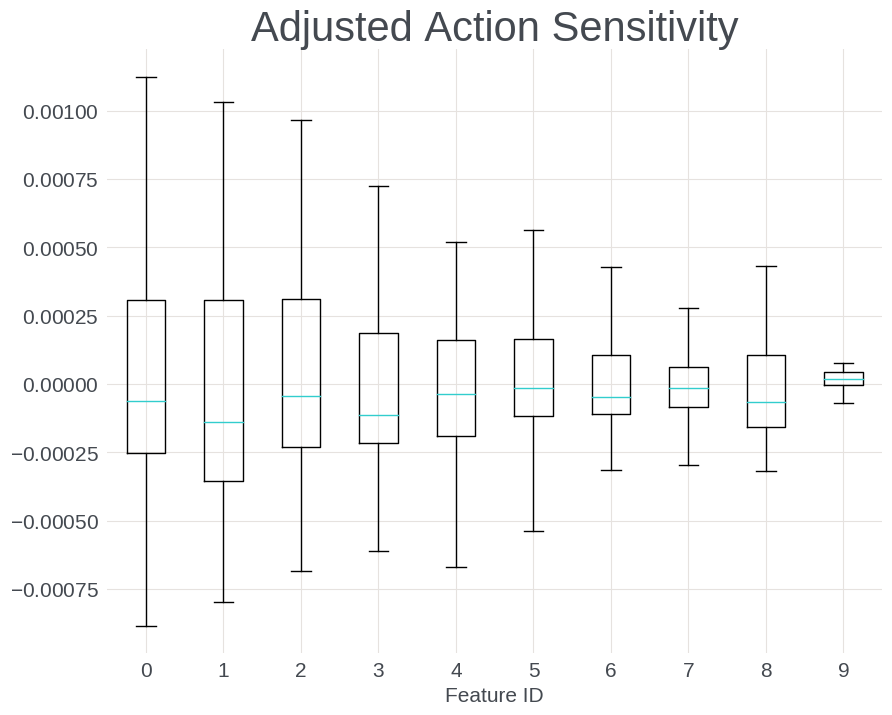}}
      \subfigure[]{\includegraphics[width=0.3\linewidth]{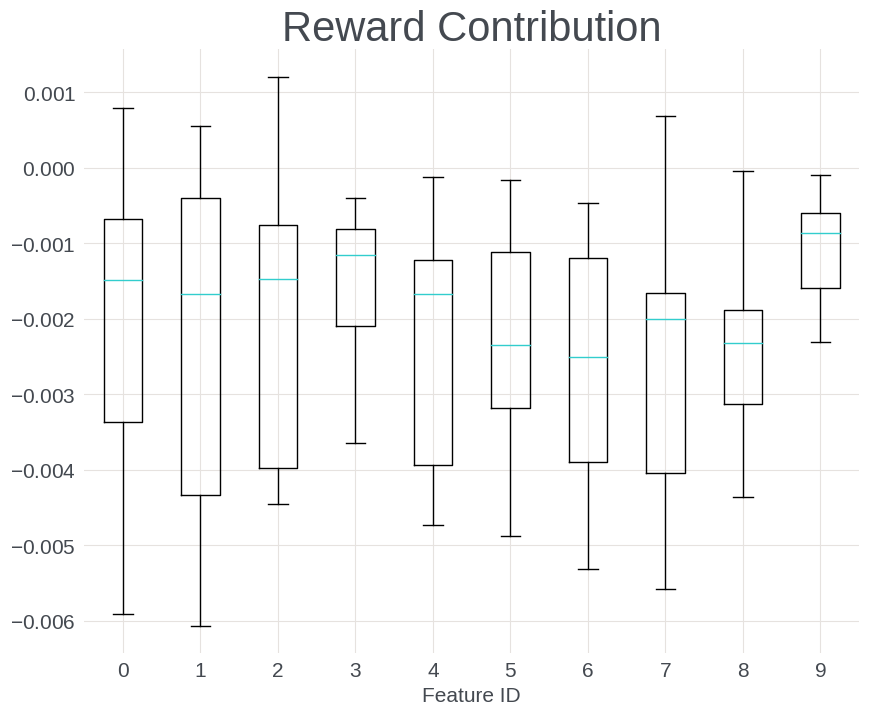}}
            \subfigure[]{


\includegraphics[width=0.2\linewidth]{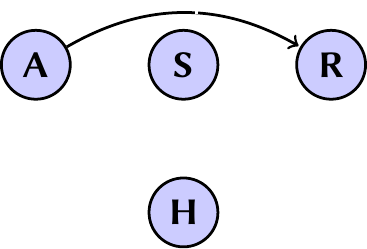}
      }
      \subfigure[]{\includegraphics[width=0.3\linewidth]{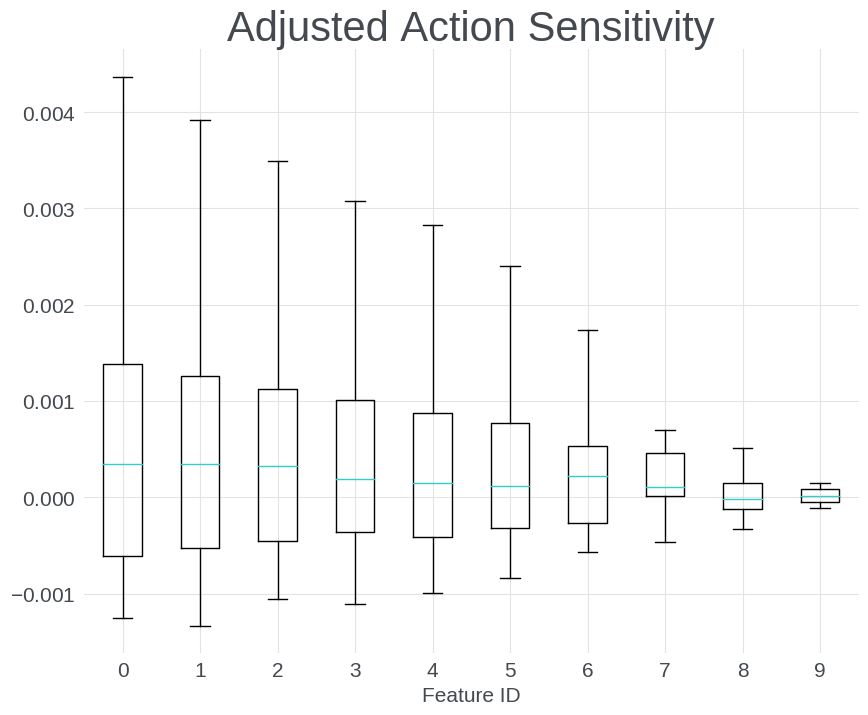}}
      \subfigure[]{\includegraphics[width=0.3\linewidth]{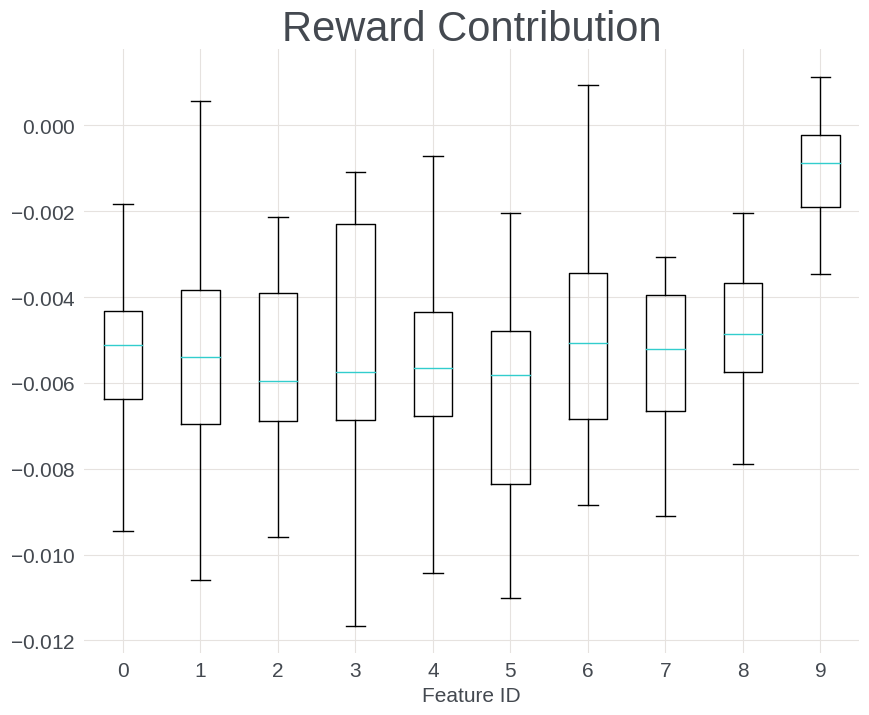}}
      \subfigure[]{


\includegraphics[width=0.2\linewidth]{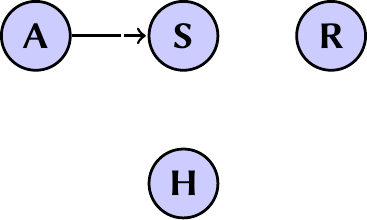}
      }
      \subfigure[]{\includegraphics[width=0.3\linewidth]{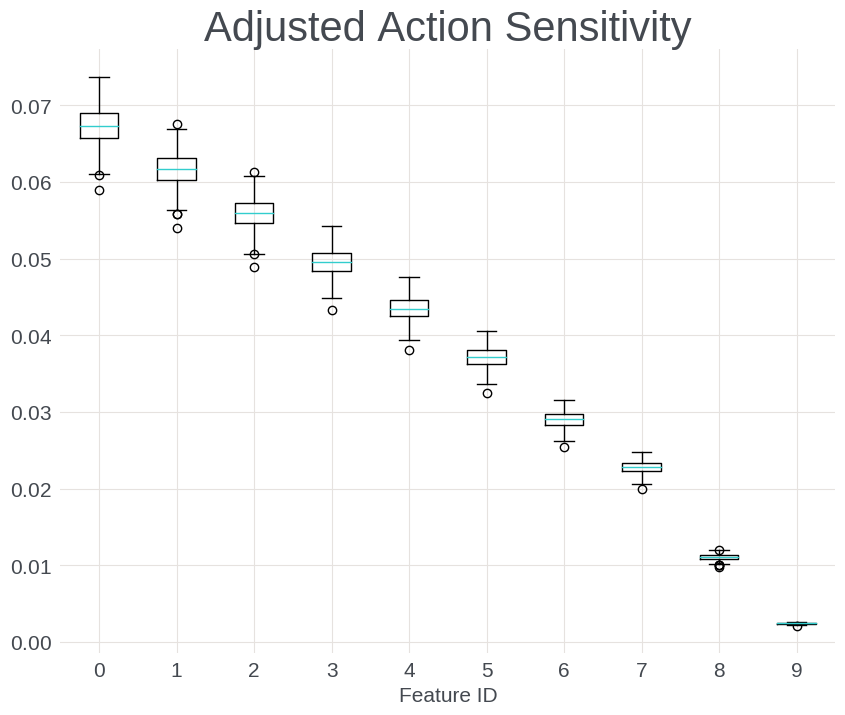}}
      \subfigure[]{\includegraphics[width=0.3\linewidth]{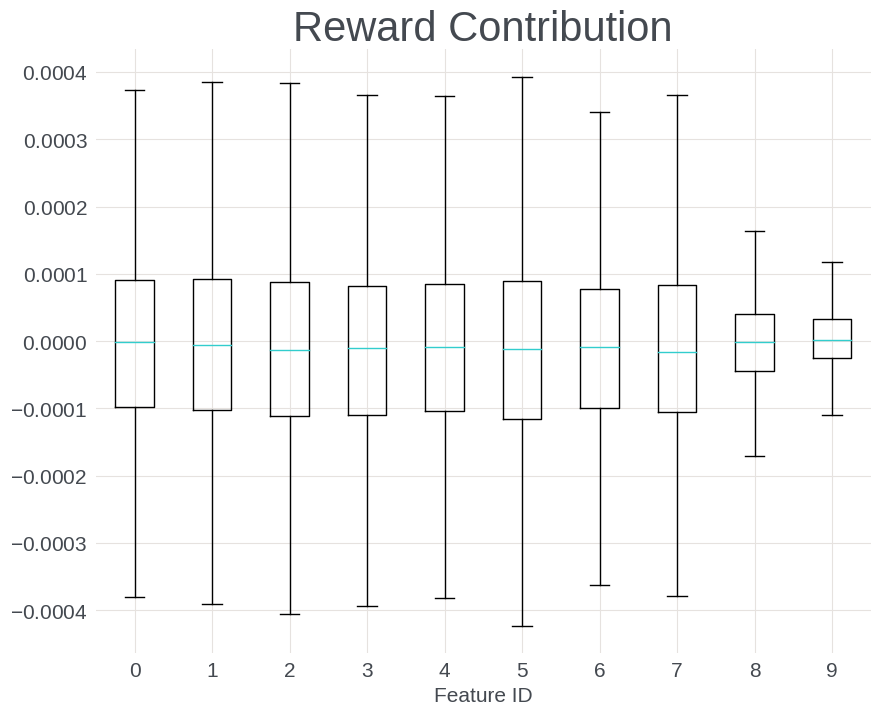}}        
    \subfigure[]{


\includegraphics[width=0.2\linewidth]{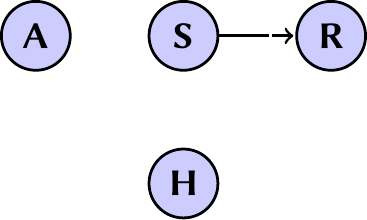}
      }
      \subfigure[]{\includegraphics[width=0.3\linewidth]{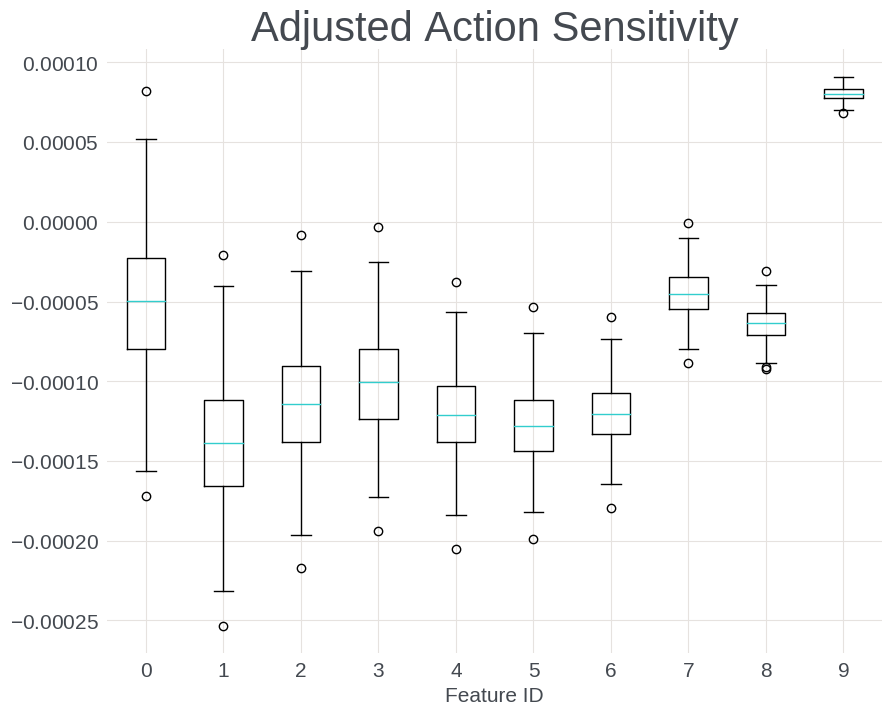}}
      \subfigure[]{\includegraphics[width=0.3\linewidth]{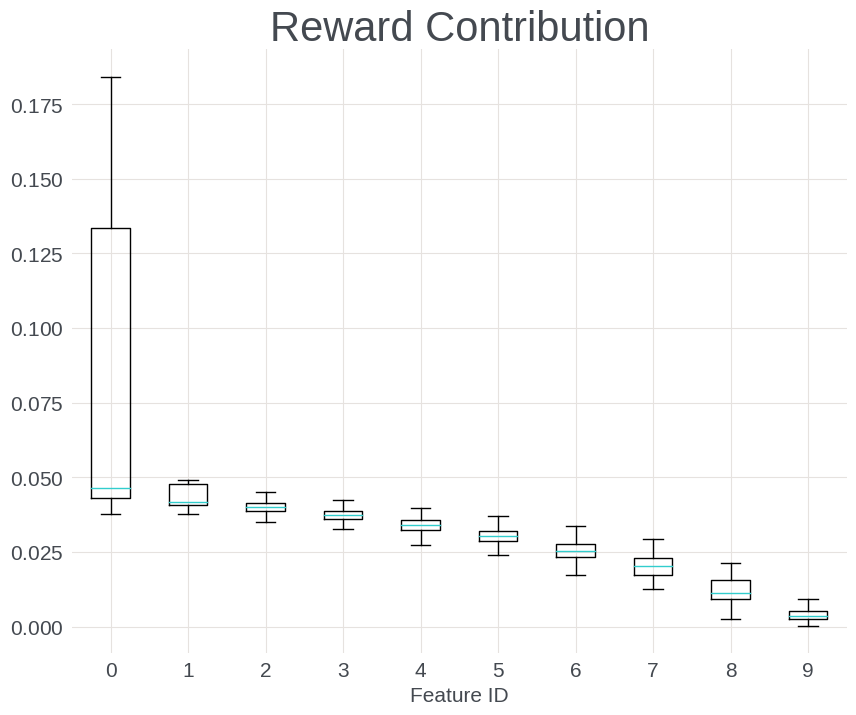}}  
      \label{fig:1}
    \end{figure}

    \begin{figure}[h]
      \centering
      \subfigure[]{


       
\includegraphics[width=0.2\linewidth]{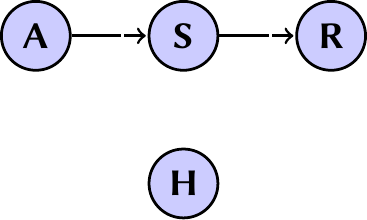}
      }
      \subfigure[]{\includegraphics[width=0.3\linewidth]{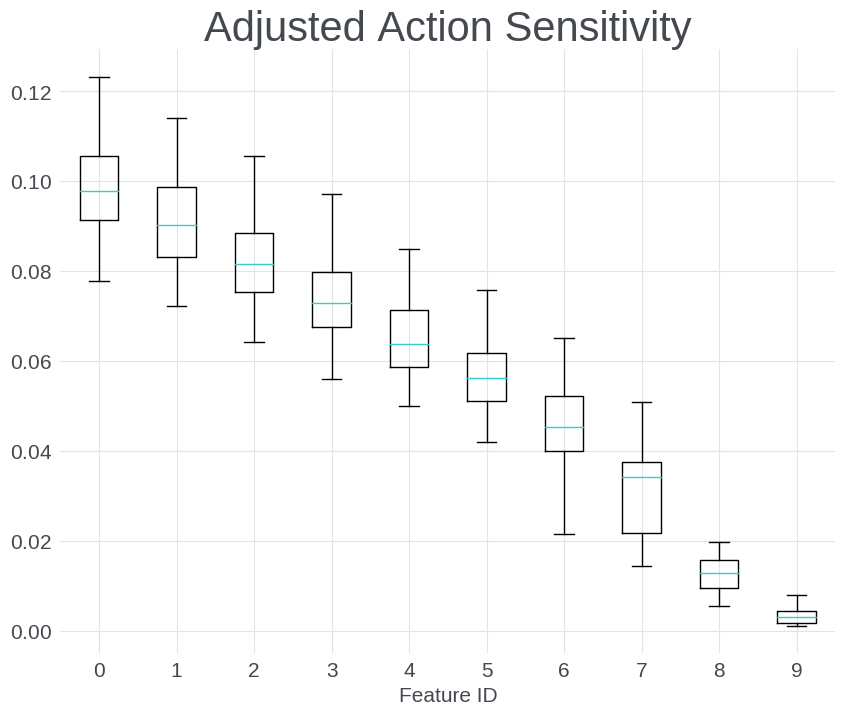}}
      \subfigure[]{\includegraphics[width=0.3\linewidth]{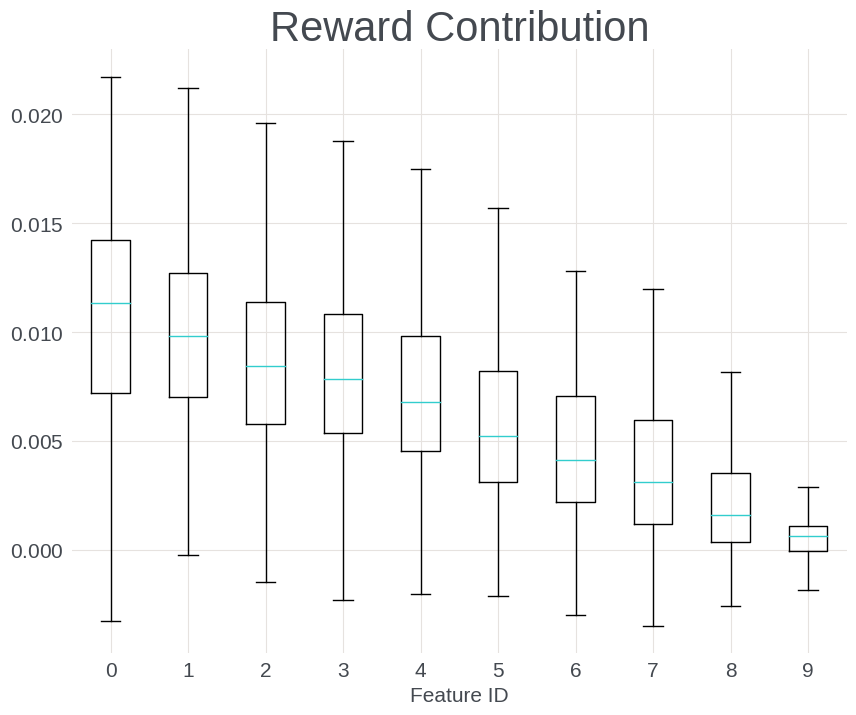}}
      \subfigure[]{


\includegraphics[width=0.2\linewidth]{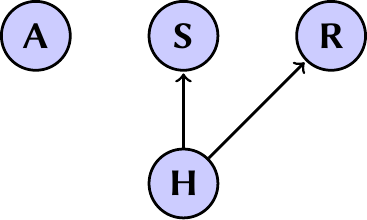}
      }
      \subfigure[]{\includegraphics[width=0.3\linewidth]{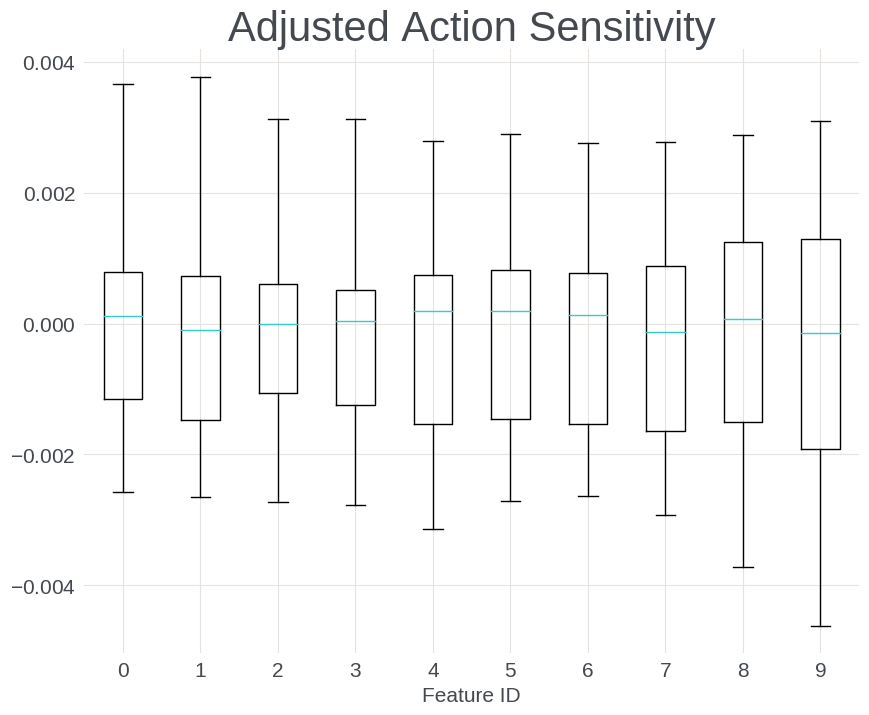}}
      \subfigure[]{\includegraphics[width=0.3\linewidth]{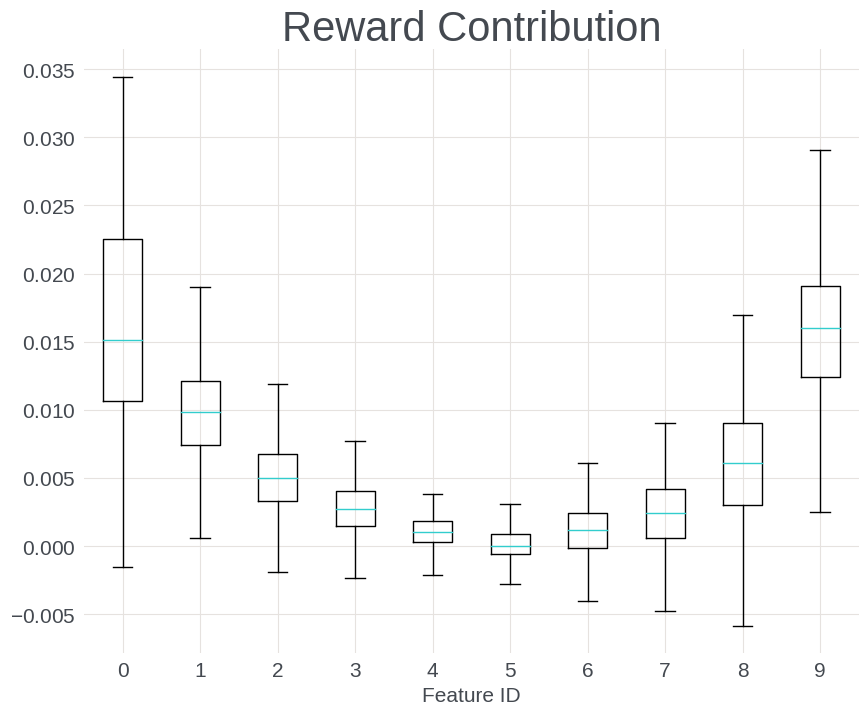}}
      \subfigure[]{


        
\includegraphics[width=0.2\linewidth]{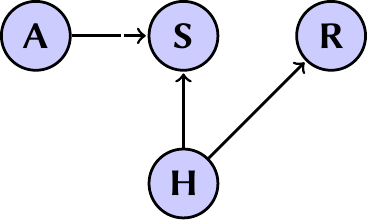}
      }
      \subfigure[]{\includegraphics[width=0.3\linewidth]{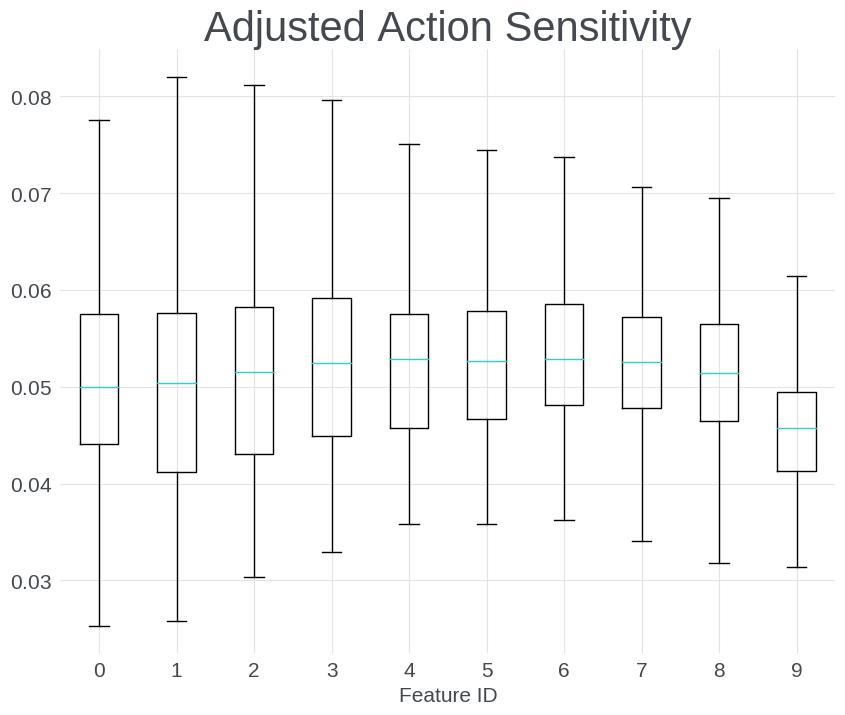}}
      \subfigure[]{\includegraphics[width=0.3\linewidth]{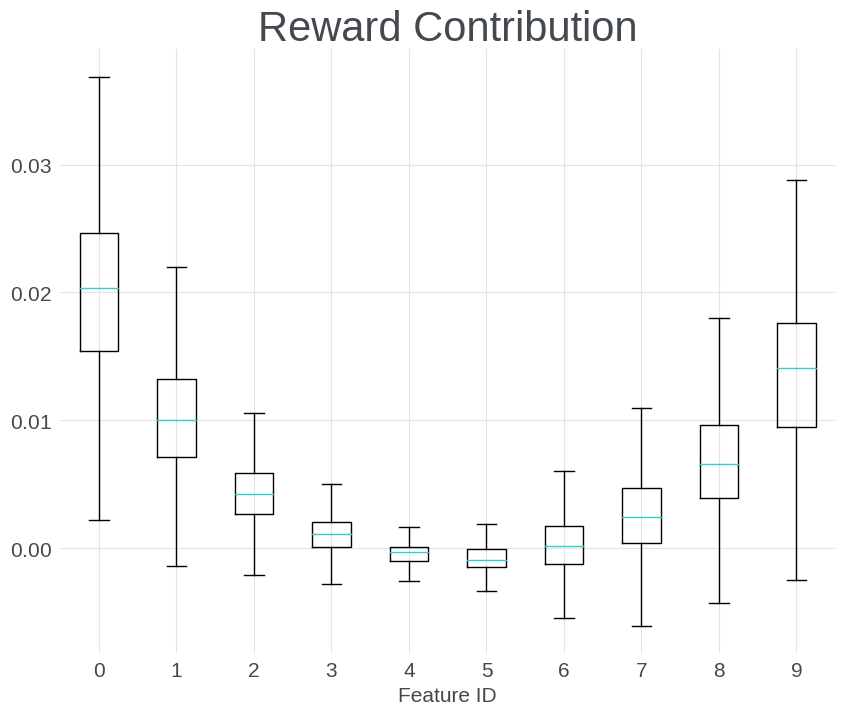}}
    \caption{Reward contribution and action sensitivity measured for different environments listed in Section~\ref{sec:env_fab}.}
      \label{result_plot}
\end{figure}

\end{document}